\documentclass[letterpaper]{article}
\usepackage{aaai}
\usepackage{graphicx}
\usepackage{times}
\usepackage{helvet}
\usepackage{courier}
\usepackage{amsmath} 
\usepackage{amssymb}  
\usepackage{amsthm} 
\usepackage{alltt}
\usepackage{xspace}
\usepackage{url}
\usepackage{html}
\usepackage[table]{xcolor}
\usepackage{color,soul}

\nocopyright

\begin{document}




\title{\textrm{The Metrics Matter!}\\\large{On the Incompatibility of Different Flavors of Replanning}}


 \author{Kartik Talamadupula$^\dag$ \and David E. Smith$^\S$ \and Subbarao Kambhampati$^\dag$
 \AND
   $^\dag$Arizona State University\\
   Dept. of Computer Science and Engg.\\   
   Tempe, AZ 85287\\
   \begin{normalsize}{\tt \{krt,rao\}} @ {\tt asu.edu}\end{normalsize} \And
     $^\S$NASA Ames Research Center\\
     Intelligent Systems Division\\
     Moffet Field, CA 94035\\
   \begin{normalsize}{\tt david.smith} @ {\tt
       nasa.gov}\end{normalsize}}


\maketitle

\begin{abstract}

  When autonomous agents are executing in the real world, the state of the
  world as well as the objectives of the agent may change from the agent's
  original model. In such cases, the agent's planning process must modify the
  plan under execution to make it amenable to the new conditions, and to
  resume execution. This brings up the \emph{replanning} problem, and the
  various techniques that have been proposed to solve it. In all, three main
  techniques -- based on three different metrics -- have been proposed in
  prior automated planning work. An open question is whether these
  metrics are interchangeable; answering this requires a normalized comparison of
  the various replanning quality metrics. In this paper, we show that it is
  possible to support such a comparison by compiling \emph{all} the respective
  techniques into a single substrate. Using this novel compilation, we
  demonstrate that these different metrics are not interchangeable, and that they
  are not good surrogates for each other. Thus we focus attention on the
  incompatibility of the various replanning flavors with each other, founded
  in the differences between the metrics that they respectively seek to
  optimize.


\end{abstract}

\section{Introduction}



Although the need for replanning has been acknowledged from the very
beginnings of automated planning research
(c.f.~\cite{fikes1972learning}), most work on replanning has
viewed it as ``technique'' rather than a problem in its own right. In
particular, most work viewed replanning from the point
of view of reducing the computational effort required to generate a new
plan, with little regard to the quality of the produced (re)plan. 
More recently, there has been some welcome effort that views replanning as a
problem rather than a technique (c.f.~\cite{cushing05,fox2006plan}). Even in such work, there has been a
significant divergence of opinion as to the right characterization of
the replanning problem. For example, while Fox {\em et. al.} argue for
plan stability as the main motivation for replanning, Cushing {\em
  et. al.} argue that sensitivity to commitments is the hallmark of
replanning. 
The fact that these multiple motivations/models for replanning persist
would seem to suggest that there is an implicit belief in the planning
community that the differences between the replanning motivations may
not be significant--and that techniques developed for one model could
well act as a good surrogates for the other models. 

In this paper we make several connected contributions on this problem. We
first show that replanning is best characterized as solving a (new) planning
problem in light of the constraints imposed by a previous plan. We will show
that the different motivations (computational efficiency, plan stability,
commitments etc.) can all be captured in this general framework, with the only
difference being in the specific form of constraints induced from the previous
plan. Second, we present a generic technique for replanning based on partial
satisfaction that is capable of simulating the different replanning
strategies. Finally, armed with this common substrate, we attempt to answer
the question: to what extent do the constraints imposed by one type of
replanning formulation act as a surrogate in tracking the constraints of
another? We do this by comparing plan stability, commitment sensitivity, and
computational efficiency across three different replanning techniques--all
implemented on the same underlying substrate. Our results show that the
different metrics are not good surrogates of each other, and lead to plans
with very different quality characteristics.


\section{Related Work}
\label{sec:relatedwork}

Replanning has been an early and integral part of automated planning and
problem solving work in AI. The STRIPS robot problem-solving system
\cite{fikes1972learning} used an execution monitoring system known as PLANEX
to recognize plan failures and replan to get back on track with the original
plan. More recent work looks at concepts such as plan {\em
stability}~\cite{fox2006plan}, which is defined as the measure of the
difference a process induces between an original plan and a new plan. This is
closely related to the idea of {\em minimal perturbation
planning}~\cite{kambhampati1990mapping} used in past replanning and plan re-use~\cite{nebel1995plan} work. Van Der Krogt \& De
Weerdt~(\citeyear{van2005plan}), on the other hand, outline a way to extend
state-of-the-art planning techniques to accommodate plan \emph{repair}. At the
other end of the spectrum, Fritz \& McIlraith~(\citeyear{fritz2007monitoring})
deal with changes to the state of the world by replanning from scratch.




Additionally, the multi-agent systems (MAS) community has also looked at
replanning issues, though more in terms of multiple agents and the conflicts
that can arise between these agents when they are executing in the same
dynamic world. Wagner et al.~(\citeyear{wagner1999multi}) proposed the twin
ideas of {\em inter-agent} and {\em intra-agent} conflict resolution. Inter-
agent commitments have been variously formalized in different work in the MAS
community
\cite{komenda2008planning,bartold2003limiting,wooldridge2000reasoning},  but
the focus has always been on the interactions between the various agents, and
how changes to the world affect the declared commitments.
Komenda et al.~(\citeyear{komenda2012decentralized}) introduce the multi-agent plan repair problem and reduce it to the multi-agent planning problem; and Meneguzzi et al.~(\citeyear{meneguzzi2013first}) introduce a first-order representation and reasoning technique for modeling commitments between agents.


\section{The Replanning Problem}

We posit that replanning should be viewed not as a technique, but as a
{\em problem} in its own right -- one that is distinct from the
classical planning problem. Formally, this idea can be stated as
follows. Consider a plan $\Pi_P$ that is synthesized in order to solve
the planning problem $P = \langle I, G \rangle$, where $I$ is the
initial state and $G$, the goal description. The world then changes
such that we now have to solve the problem $P^\prime = \langle
I^\prime, G^\prime \rangle$, where $I^\prime$ represents the changed
state of the world, and $G^\prime$ a changed set of goals (possibly
different from $G$). We then define the {\em replanning problem} as
one of finding a new plan $\Pi_P^\prime$ that solves the problem
$P^\prime$ subject to a set of constraints $\psi^{\Pi_P}$. This model
is depicted in Figure~\ref{replan-model-fig}. The composition of the
constraint set $\psi^{\Pi_P}$, and the way it is handled, can be
described in terms of specific {\em models} of this newly formulated
replanning problem. Here, we present three such models based on the
manner in which the set $\psi^{\Pi_P}$ is populated.

\begin{enumerate}

\item {\bf $M_1$ $|$ Replanning as Restart}: This model treats
  replanning as `planning from restart' -- i.e., given changes in the
  world $P = \langle I, G \rangle \rightarrow\ P^\prime
  =$\nolinebreak[4] $\langle I^\prime, G^\prime \rangle$, the old plan
  $\Pi_P$ is completely abandoned in favor of a new plan $\Pi_P^\prime$
  which solves $P^\prime$. Thus the previous plan induces no
  constraints that must be respected, meaning that the set
  $\psi^{\Pi_P}$ is empty.

  \begin{figure}[htp]
    \centering
    \includegraphics[keepaspectratio, width=225pt, clip]
    {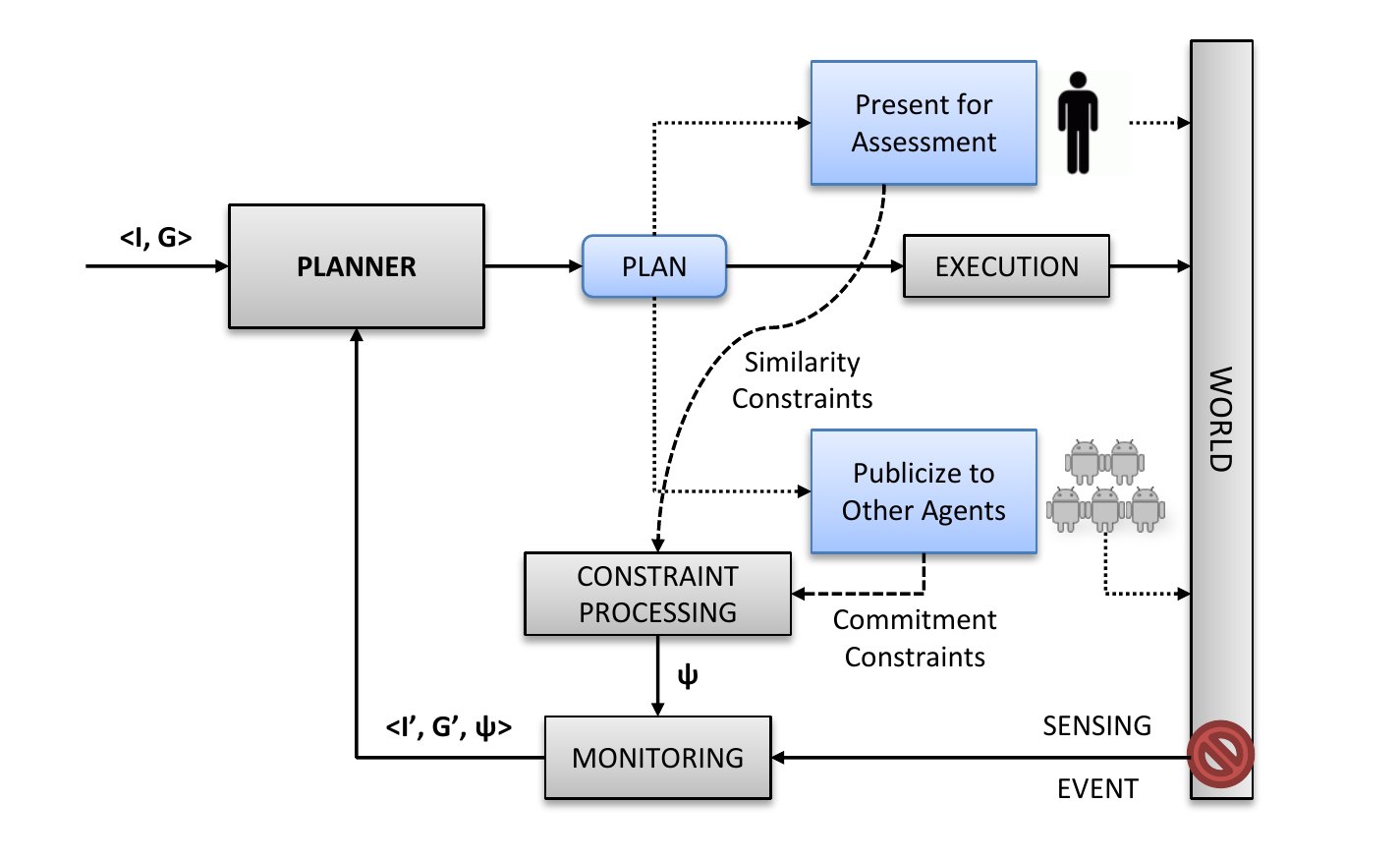}
    \caption{A model of replanning}
    \label{replan-model-fig}
  \end{figure}

\item {\bf $M_2$ $|$ Replanning to Reduce Verification Cost}: When the state
  of the world forces a change from a plan $\Pi_P$ to a new one
  $\Pi_P^\prime$, in the extreme case, $\Pi_P^\prime$ may bear no
  relation to $\Pi_P$. However, it may be desirable that the cost
  of comparing the differences between the two
  plans with respect to execution in the world be reduced as far as
  possible (we explore all the possible reasons for this in Section~\ref{subsec:replan_efficiency}). The problem of minimizing this cost can be re-cast as one
  of minimizing the differences between the two plans $\Pi_P^\prime$
  and $\Pi_{P}$ using {\em syntactic constraints} on the form of the
  new plan. These syntactic constraints are added to the set $\psi^{\Pi_P}$.

\item {\bf $M_3$ $|$ Replanning to Respect Commitments}: In many
  real world scenarios, there are multiple agents $A_1 \ldots A_n$
  that share an environment and hence a world state.\footnote{Note
    that this is the case regardless of whether the planner models
    these agents explicitly or chooses to implicitly model them in the
    form of a dynamic world.} The individual plans of these agents,
  $\Pi_1 \ldots \Pi_n$ respectively, affect the common world state
  that the agents share and must plan in.  This leads to the formation
  of dependencies, or {\em commitments}, by other agents on an agent's
  plan. These commitments can be seen as special types of \emph{soft} constraints
  that are induced by an executing plan; they come with a penalty that is assessed when a given commitment constraint is not satisfied by the replan. The aggregation of
  these commitments forms the set $\psi^{\Pi_P}$ for this model.

\end{enumerate}

\noindent In the following section, we explore the composition of the
constraint set $\psi^{\Pi}$ (for any given plan $\Pi$) in more detail.

\section{Replanning Constraints}
\label{sec:replanning-constraints}

As outlined in the previous section, the replanning problem can be
decomposed into various {\em models} that are defined by the
constraints that must be respected while transitioning from the old
plan $\Pi$ to the new plan $\Pi^\prime$. In this section, we define
those constraints, and explore the composition of the set
$\psi$ for each of the models defined previously.

\subsection{Replanning as Restart}

By the definition of this model, the old plan $\Pi_P$ is completely
abandoned in favor of a new one. There are no constraints induced by
the previous plan that must be respected, and thus the set
$\psi^{\Pi_P}$ is empty.

\subsection{Replanning to Reduce Verification Cost}
\label{subsec:replan_efficiency}

It is often desirable that the replan for the new problem instance
$P^\prime$ resemble the previous plan $\Pi_P$ in order to reduce the
computational effort associated with verifying that it still meets the
objectives, and to ensure that it can be carried out in the world. We
name the effort expended in this endeavor as the {\em reverification
  complexity} associated with a pair of plans $\Pi_P$ and
$\Pi_P^\prime$, and informally define it as the amount of effort/computation that
an agent has to expend on comparing the differences between an old
plan $\Pi_P$ and a new candidate plan $\Pi_P^\prime$ with respect to
executability in the world.


Such comparison may be necessitated due to one of three reasons:

\begin{itemize}
  
  \item $R1$ - \textbf{Communication}: Changes may have to be communicated to other agents, which may have predicated their own plans on actions in the original plan. 

  \item $R2$ - \textbf{Explanation}: Changes may have to be \emph{explained} to other agents, like human supervisors.

  \item $R3$ - \textbf{Incomplete Models}: Additional simulations or computation may have to occur to make up for a plan that was made with an incomplete model of the world.

\end{itemize}

\noindent Real world examples where reverification complexity is of utmost
importance abound, including machine-shop or factory-floor planning, planning
for assistive robots and teaming, and planetary rovers. Past work on replanning has addressed this
problem via the idea of {\em plan stability}~\cite{fox2006plan}. The general
idea behind this approach is to preserve the stability of the replan
$\Pi_P^\prime$ by minimizing some notion of difference with the original plan
$\Pi_P$. In the following, we examine two such ways of measuring the
difference between pairs of plans, and how these can contribute constraints to
the set $\psi^{\Pi_P}$ that will minimize reverification complexity.

\subsubsection{Action Similarity}
\label{subsubsec:action-sim}

The most obvious way to compute the difference between a given pair of plans
is to compare the actions that make up those plans. Fox et
al.~(\citeyear{fox2006plan}) define a way of doing this - given an original
plan $\Pi$ and a new plan $\Pi^\prime$, they define the difference between
those plans as the number of actions that appear in $\Pi$ and not in
$\Pi^\prime$ plus the number of actions that appear in $\Pi^\prime$ and not in
$\Pi$. If the plans $\Pi$ and $\Pi^\prime$ are seen as sets comprised of
actions, then this is essentially the symmetric difference of those sets, and
we have the following constraint:\footnote{A different measure of similarity considers the
\emph{causal links} in a plan; space considerations preclude a discussion of
this measure.} $\min\ | \Pi\ \triangle \ \Pi^\prime |$. This measure corresponds to $R2$ above, since any deviations from the original plan -- either in terms of additions or deletions -- have to be explained.

Yet another way of computing whether $\Pi^\prime$ is a good replan is to
determine how many of the actions in $\Pi$ are retained by the new plan. To
compute this value, both plans can be seen as sets of actions, and the set
difference of $\Pi$ and $\Pi^\prime$ must then be minimized; this gives us the
following constraint: $\min | \Pi\ \setminus \ \Pi^\prime |$. In our
compilation (detailed in
Section~\ref{subsec:compiling_action_similarity_to_psp}), we use the set
difference constraint as the metric to be minimized; however, we report both
the set and symmetric difference values in our evaluation. This metric
corresponds to $R1$ give above, since only actions that were part of the
original plan but deleted from the replan have to be communicated to other
agents.

\subsection{Replanning to Respect Commitments}
\label{subsec:replan_commitments}

\begin{quote} {\em In a multiperson situation, one man's goals may be\\ 
    another man's constraints.} -- Herb Simon~(\citeyear{simon1964concept})
\end{quote}

\noindent In an ideal world, a given agent would be the sole center of plan
synthesis as well as execution, and replanning would be necessitated only by
those changes to the world state that it cannot foresee. However, in the real
world, there exist multiple such agents, each with their own disparate
objectives but all bound together by the world that they share. A plan $\Pi_P$
that is made by a particular agent affects the state of the world and hence
the conditions under which the other agents must plan -- this is true in turn
for every agent. In addition, the publication of a plan $\Pi^A_P$ by an agent
$A$ leads to other agents predicating the success of their own plans on parts
of $\Pi^A_P$, and complex dependencies are developed as a result. Full multi-agent planning can resolve the issues that arise out of changing plans in such
cases, but it is far from a scalable solution for real world domains
currently. Instead, this multi-agent space filled with dependencies can be
projected down into a single-agent space with the help of {\em commitments} as
defined by Cushing \& Kambhampati~(\citeyear{cushing05}). These commitments are related to an agent's
current plan $\Pi$, and can describe different requirements that come about:
(i) when the agent decides to execute $\Pi$, and other agents predicate their
own plans on certain aspects of it; (ii) due to cost or time based
restrictions imposed on the agent; or (iii) due to the agent having paid an up-front setup cost to enable some part of the plan $\Pi$.






It may seem as though the same kinds of constraints that seek to minimize
reverification complexity between plans $\Pi$ and $\Pi^\prime$ (minimizing
action and causal link difference between plans) will also serve to preserve
and keep the most commitments in the world. Indeed, in extreme cases, it might
even be that keeping the structures of $\Pi$ and $\Pi^\prime$ as similar as
possible helps keep the maximum number of commitments made due to $\Pi$.
However, this is certainly not the most natural way of keeping commitments. In
particular, this method can fail when there is any significant deviation in
structure from $\Pi$ to $\Pi^\prime$; unfortunately, most unexpected changes
in real world scenarios are of a nature that precludes retaining significant
portions of the previous plan. Instead, we model commitments as state
conditions, and the constraints that mandate the preservation of commitments
as \emph{soft goals} that the planner seeks to satisfy. We elaborate on this
in Section~\ref{subsec:psp_technique}.

\section{Compilation to a Single Substrate} 
\label{sec:compilation_to_a_single_substrate}
\label{sec:solution_techniques}


Both kinds of constraints discussed in the previous section -- dealing with
plan similarity, as well as with inter-agent commitments -- can be cast into a
single planning substrate. In this section, we first demonstrate compilations
from action similarity and inter-agent commitments to partial satisfaction
planning (PSP). We then detail a simple compilation from PSP to preference-based planning (PBP).

\subsection{I: Partial Satisfaction Planning}
\label{subsec:psp_technique}

We follow~\cite{vandenbriel2004eap} in defining a PSP net benefit problem
as a planning problem $P = (F, O, I, G_s)$, where $F$ is a finite set
of fluents, $O$ is a finite set of operators, and $I \subseteq F$ is
the initial state as defined earlier in our paper. For each goal $g \in G$ from the original set of goals, a soft goal $g_s$ with
a penalty $p_g$ is created; the set of all soft goals
thus created is added to a new set $G_s$.

The intuition behind casting replanning constraints as goals is that a new
plan (replan) must be constrained in some way towards being similar to
the earlier plan. However, making these `replan constraint goals' {\em hard} would
over-constrain the problem -- the change in the world from $I$ to
$I^\prime$ may have rendered some of the earlier actions, or commitments, impossible to preserve. Therefore the replanning constraints
are instead cast as {\em soft} goals, with penalties that are assessed when they are violated. In order to support the action similarity or inter-agent commitment preservation goals, new fluents
need to be added to the domain description that indicate the execution
of an action or achievement of a fluent respectively. Further, new
copies of the existing actions in the domain must be added to house
these effects.


\subsubsection{Compiling Action Similarity to PSP}
\label{subsec:experimental_setup}
\label{subsec:compiling_action_similarity_to_psp}

The first step in the compilation is converting the action similarity
constraints in $\psi^{\Pi^A_P}$ to soft goals to be added to $G_s$.
Before this, we examine the structure of the constraint set
$\psi^{\Pi^A_P}$; for every ground action $\bar{a}$ (with the names of
the objects that parameterize it) in the old plan $\Pi$, the
corresponding action similarity constraint is $\Psi_{\bar{a}} \in
\psi^{\Pi^A_P}$, and that constraint stores the name of the action as
well as the objects that parameterize it.



Next, a copy of the set of operators $O$ is created and named
$O_{as}$; similarly, a copy of $F$ is created and named $F_{as}$. For
each (lifted) action $a \in O_{as}$ that has an instance in the
original plan $\Pi$, a new fluent named ``$a$-executed'' (along with
all the parameters of $a$) is added to the fluent set $F_{as}$.  For each action $a \in O_{as}$, a new action $a_{as}$ -- which is a copy
of $a$ that additionally also gives the predicate
$a$-executed as an effect -- is created. In the worst case, the number of
actions in each $O_{as}$ could be twice the number in $O$.


Finally, for each constraint $\Psi_{\bar{a}} \in \psi^{\Pi^A_P}$, a
new soft goal $g_{\bar{a}}$ is created with corresponding penalty values $p_{g_{\bar{a}}}$,
and the predicate used in $g_{\bar{a}}$ is $\bar{a}$-executed
(parameterized with the same objects that $\bar{a}$ contains) from
$O_{as}$. All the $g_{\bar{a}}$ goals thus created are added to $G_s$.
In order to obtain the new compiled replanning instance $P^\prime$
from $P$, the initial state $I$ is replaced with the state at which
execution was terminated, $I^\prime$; the set of operators $O$ is
replaced with $O_{as}$; and the set of fluents $F$ is replaced with
$F_{as}$. The new instance $P^\prime~=~(F_{as}, O_{as}, I^\prime,
G_s)$ is given to a PSP planner to solve.

\subsubsection{Compiling Commitments to PSP} 
\label{ssub:compiling_commitments_to_psp}




Inter-agent commitments can be compiled to PSP in a manner that is very
similar to the above compilation. The difference that now needs to be
considered is that the constraints are no longer on actions, but on the
grounded fluents that comprise the commitments in a plan instead.

The first step is to augment the set of fluents; a copy of $F$ is created and
named $F_{cs}$.  For every fluent that is relevant to the inter-agent
commitments $f \in F$ (an example of such fluents is provided in
Section~\ref{ssub:agent_commitments}), a new fluent named ``$f$-achieved'' is
added to $F_{cs}$, along with all the original parameters of $f$. A copy of
the set of operators $O$ is created and named $O_{cs}$. Then, for each action
$a \in O_{cs}$, a new action $a_{cs}$ is added; $a_{cs}$ is a copy of the
action $a$, with the additional effects that for every commitment-relevant
fluent $f_a$ that is in the add effects of the original $a$, $a_{cs}$ contains
the effect {$f_a$-achieved}. 

Finally, the commitment constraints in $\psi^{\Pi^A_P}$ must be converted to
soft goals that can be added to $G_s$. The constraints $\Psi \in
\psi^{\Pi^A_P}$ are obtained by simulating the execution of $\Pi$ from $I$
using the operators in $O$. Each ground commitment-relevant effect $\bar{f_e}$
of a commitment-relevant action $\bar{a_\Pi}$ in $\Pi$ is added as a new
constraint $\Psi_{\bar{f_e}}$. Correspondingly, for each such new constraint
added, a new soft goal $g_{\bar{f_e}}$ is created whose fluent corresponds to
$\bar{f_e}$, with penalty value $p_{g_{\bar{f_e}}}$. All the goals thus created are added to
$G_s$. The new planning instance to be provided to the PSP planner is thus
given as $P^\prime = (F_{cs}, O_{cs}, I^\prime, G_s)$.

\subsection{II: Preferences} 
\label{sub:t4_preferences}



The constraints in the set $\psi^{\Pi^A_P}$ can also be cast as
\emph{preferences}~\cite{jorge2009planning} on the new plan that needs to be
generated by the replanning process. Preferences are indicators of the
\emph{quality} of plans, and can be used to distinguish between plans that all
achieve the same goals. The automated planning community has seen a lot of
work in recent years on fast planners that solve preference-based planning
problems specified using the PDDL3~\cite{gerevini2006plan} language; casting
the constraints in $\psi^{\Pi^A_P}$ into preferences can thus open up the use
of these state-of-the-art planners in solving the replanning problem. Benton
et al.~(\citeyear{benton09}) have already detailed a compilation that
translates simple preferences specified in PDDL3 to soft
goals. This work can be used in order to translate the replanning constraints
into simple preferences, thus enabling the use of planners like
SGPlan5~\cite{hsu2007constraint} and OPTIC~\cite{benton2012temporal}. In our
evaluation, we use this preference-based approach to improve the scalability
of our result generation.

The compilation itself is straightforward. For every soft goal $g_s$ that
models either an action similarity or inter-agent commitment constraint
respectively (from Section~\ref{subsec:psp_technique}), we create a new preference $\tau_s$ where the condition that is
evaluated by the preference is the predicate $a$-executed or $f$-achieved
respectively, and the penalty for violating that commitment is the penalty
value associated with the soft goal $p_{g_s}$. The set of preferences thus
created is added to the problem instance, and the metric is set to minimize
the (unweighted) sum of the preference violation values.

\section{Evaluation}






The compilation outlined in Section~\ref{sec:solution_techniques} serves as
support for our first claim -- that it is possible to support all the existing
replanning metrics (and associated techniques) using a single planner, via
compilation to a single substrate. That substrate can be either soft goals
(and the technique to solve them partial satisfaction planning), or
preferences (preference-based planning). In this section, we provide empirical
support for our second point -- namely that these different replanning metrics
are not good surrogates for each other -- and that swapping them results in a
deterioration of the metric being optimized.





\subsection{The Warehouses Domain}
\label{sec:domain_warehouses}

Planning for the operations and agents contained in automated warehouses has emerged as an important application, particularly with the success of
large-scale retailers like Amazon. Given the size, complexity, as well as
real-time nature of the logistical operations involved in administering and
maintaining these warehouses, automation is inevitable. One motivation behind designing an entirely new domain\footnote{We plan to release this domain to the planning community for testing purposes.} for our evaluations was so that we
could control the various actions, agents, and problem instances that were
generated. Briefly, our domain consists of \emph{packages} that are originally
stocked on \emph{shelves}; these shelves are accessible only from certain
special locations or \emph{gridsquares}. The gridsquares are themselves
connected in random patterns to each other (while ensuring that there are no
isolated gridsquares). \emph{Carriers} -- in the form of \emph{forklifts} that
can stock and unstock packages from shelves, and \emph{transports} that can
transport packages placed on them between various gridsquares -- are used to
shift the packages from their initial locations on shelves to
\emph{packagers}, where they are packaged. The instance goals are all
specified in terms of packages that need to be packaged.

\subsubsection{Perturbations} 
\label{ssub:perturbations}

There are \emph{two} main kinds of perturbations that we model and generate: (i)
packages can \emph{fall off} their carriers at random gridsquares; and (ii)
carriers (forklifts or transports) can themselves \emph{break down} at random.
For packages that fall off at a gridsquare, a forklift is required at that
gridsquare in order to lift that package and transport it to some other
desired location (using either that same forklift, or by handing off to some
other carrier). For carriers that break down, the domain contains special
\emph{tow-trucks} that can attach themselves to the carrier and tow it along
to a \emph{garage} for a repair action to be performed. Garages are only
located at specific gridsquares.


\subsubsection{Agent Commitments} 
\label{ssub:agent_commitments}

There are three kinds of agents in our domain -- packagers, tow-trucks, and
carriers. Agent commitments are thus any predicates that these agents
participate in (as part of the state trace of a given plan $\Pi$). In our
domain, there are four such predicates: forklifts \emph{holding} packages,
packages \emph{on} transports, tow-trucks \emph{towing} carriers, and packages
\emph{delivered} to a packager.

\subsection{Results} 
\label{sub:results}


\subsubsection{Experimental Setup} 
\label{ssub:experimental_setup}


Using the domain described in Section~\ref{sec:domain_warehouses}, we created
an automated problem generator that can generate problem instances of
increasing complexity. Instance complexity was determined by the number of
packages that had to be packaged, and ranged from 1 to 12.\footnote{The
objective of this paper is not to demonstrate the scalability of either the
planner or the domain in question, but rather to show the difference in
performance when different replanning metrics are substituted for each other.}
We associated four randomly generated instances with each step up in
complexity, for a total of 48 problem instances. As the number of packages
increased, so did the number of other objects in the instance -- forklifts,
transports, shelves, and gridsquares. The number of tow-trucks and garages was
held constant at one each per instance. The initial configuration of all the
objects (through the associated predicates) was generated at random, while the top-level goals were always to have packaged all the packages.


For each of the replanning metrics that we are interested in evaluating --
speed, similarity, and commitment satisfaction -- we set up the constraints
outlined in Section~\ref{sec:replanning-constraints} as part of the replanning
metric. When optimizing the time taken to generate a new plan, the planner
does not need to model any new constraints, and can choose any plan that is
executable in the changed state of the world. Likewise, when the planner is
optimizing the similarity between the new plan and the previous plan (as
outlined in Section~\ref{subsubsec:action-sim}), it only evaluates the number
of differences (in terms of action labels) between the two plans, and chooses
the one that minimizes that value. The planner's search is directed towards
plans that fulfill this requirement via the addition of \emph{similarity
goals} to the existing goal set, via the compilation procedure described in
Section~\ref{subsec:experimental_setup}. Finally, when optimizing the
satisfaction of commitments created by the old plan that must be satisfied by
the new one, the planner merely keeps track of how many of these are
fulfilled, and ranks potential replans according to that. These commitments
are added as additional (simple) preferences to the planner's goal set, and in our current evaluation each preference has the same violation cost (1 unit) associated with it.


All the problem instances thus generated were solved with the SGPlan5
planner~\cite{hsu2007constraint}, which handles preference-based planning
problems via partition techniques by using the costs associated with violating
preferences to evaluate partial plans. The planner was run on a virtual
machine on the Windows Azure A7 cluster featuring eight 2.1 GHz AMD Opteron
4171 HE processors and 56GB of RAM, running Ubuntu 12.04.3 LTS. All the
instances were given a 90 minute timeout; instances that timed out do not have
data points associated with them.

\begin{figure}[h]
  \centering
  \includegraphics[keepaspectratio, width=250pt, clip]
  {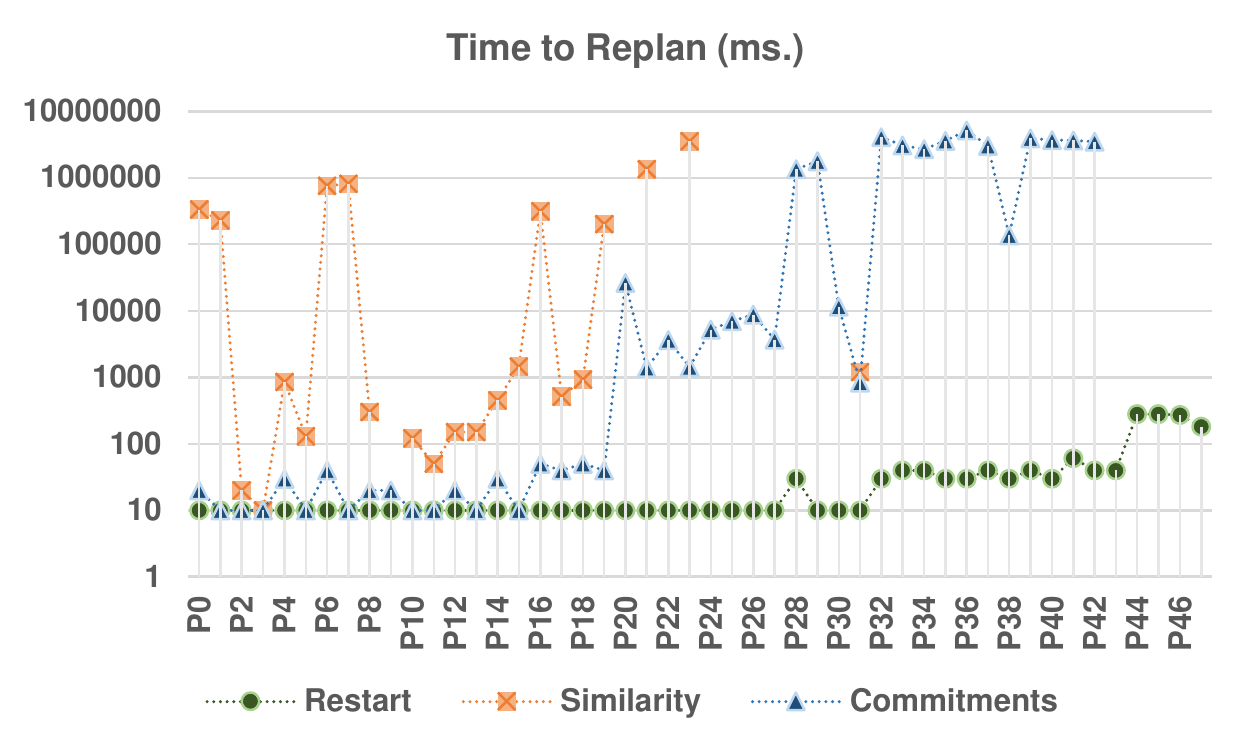}
  \vspace{-8mm}
  \caption{Time taken to replan, in milliseconds (ms.)}
  \label{replan-time}
\vspace{-3mm}
\end{figure}

\begin{figure}[h]
  \centering
  \includegraphics[keepaspectratio, width=250pt, clip]
  {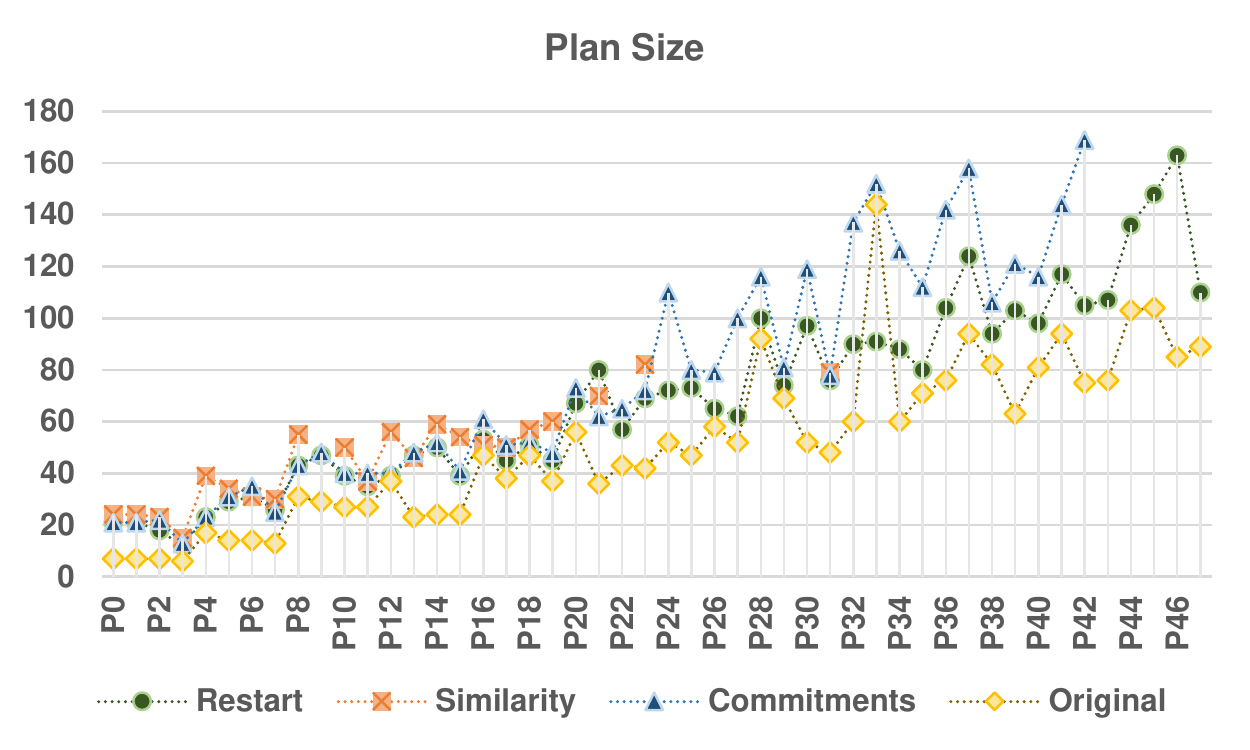}
  \vspace{-8mm}
  \caption{Plan size (number of actions)}
  \label{plan-size}
\vspace{-3mm}
\end{figure}





\subsubsection{Metric: Speed} 


In Figure~\ref{replan-time}, we present the time taken for the planner to
generate a plan (on a logarithmic scale) for the respective instances, using
the three replanning constraint sets. Replanning as restart is a clear winner,
since it takes orders of magnitude less time than the other two methods to
come up with a plan. In particular, replanning that takes plan similarity into
account takes an inordinate amount of time in coming up with new plans, even
for the smaller problem instances. This shows that when speed is the metric
under consideration, neither similarity with the original plan nor respecting
the inter-agent commitments are good surrogates for optimizing that metric. It
must be pointed out here that our method of evaluation does not re-use any of
the search effort while generating the replan; however, the findings of Nebel
\& Koehler~(\citeyear{nebel1995plan}) ensure that this is not a concern.

Additionally, we also measured the length of the plans that were generated, in
order to compare against the original plan length. Figure~\ref{plan-size}
shows that the planner doesn't necessarily come up with significantly
\emph{longer} plans when it has to replan; instead, most of the computation
time seems to be spent on optimizing the metric in question. However, these
results seem to indicate that if plan \emph{length} is the metric that is
sought to be optimized, replanning without additional constraints (as restart)
is the way to go.



\begin{figure}[h]
  \centering
  \includegraphics[keepaspectratio, width=250pt, clip]
  {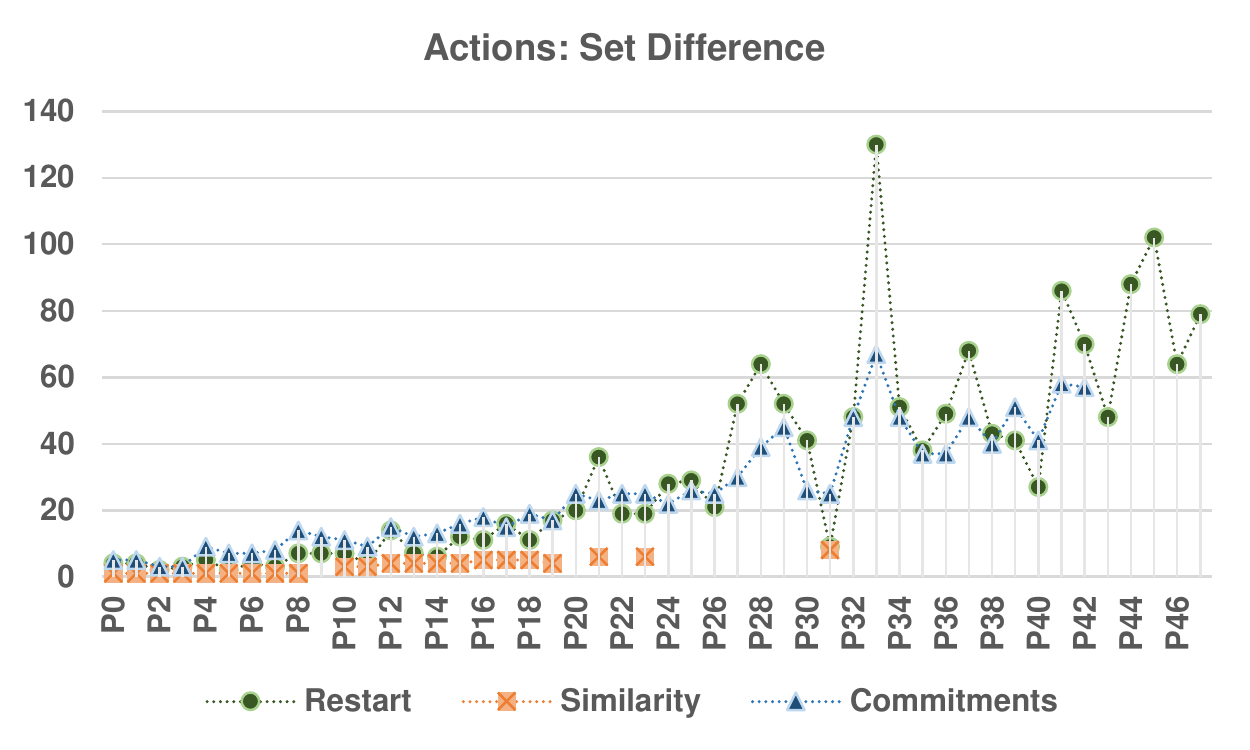}
  \vspace{-8mm}
  \caption{Set difference (action) vs. original plan $\Pi$}
  \label{set-diff}
\vspace{-3mm}
\end{figure}

\begin{figure}[h]
  \centering
  \includegraphics[keepaspectratio, width=250pt, clip]
  {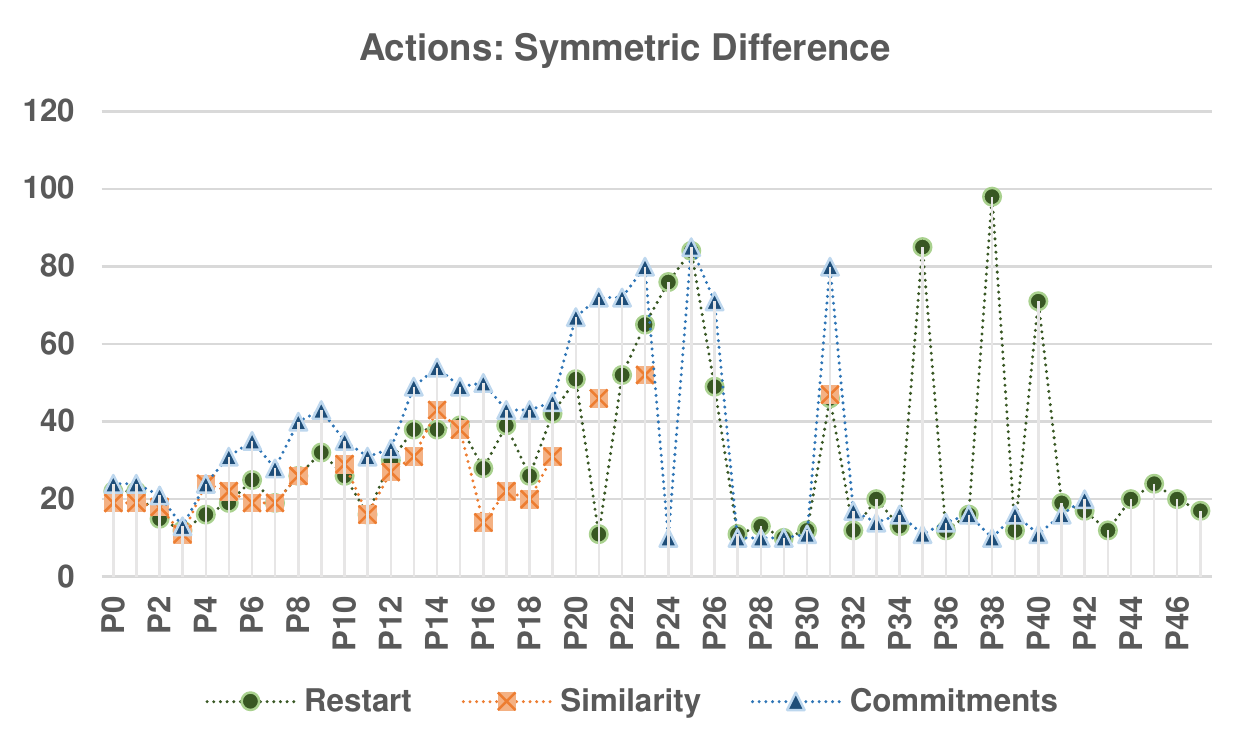}
  \vspace{-8mm}
  \caption{Symmetric difference (action) vs. original plan $\Pi$}
  \label{sym-diff}
\vspace{-3mm}
\end{figure}

\subsubsection{Metric: Similarity} 

For this evaluation, we modeled the difference between the old plan $\Pi$ and
the new replan $\Pi^\prime$ as the set difference $| \Pi\ \setminus \
\Pi^\prime |$ between the respective action sets. We then plotted this number
for the different problem instances as a measure of the differences between
the two plans. As shown in Figure~\ref{set-diff}, the method that takes plan
similarity constraints into consideration does much better than the other two
for this case. Additionally, we also calculated the symmetric difference
$| \Pi\ \triangle \ \Pi^\prime |$ (the metric used by Fox et
al.~\cite{fox2006plan}); these results are presented in Figure~\ref{sym-diff}.
Even here, the approach that respects the similarity constraints does
consistently better than the other two approaches. Thus these two results show
that when similarity with the original plan is the metric to be maximized,
neither of the other two methods can be used for quality optimization.


\begin{figure}[h]
  \centering
  \includegraphics[keepaspectratio, width=250pt, clip]
  {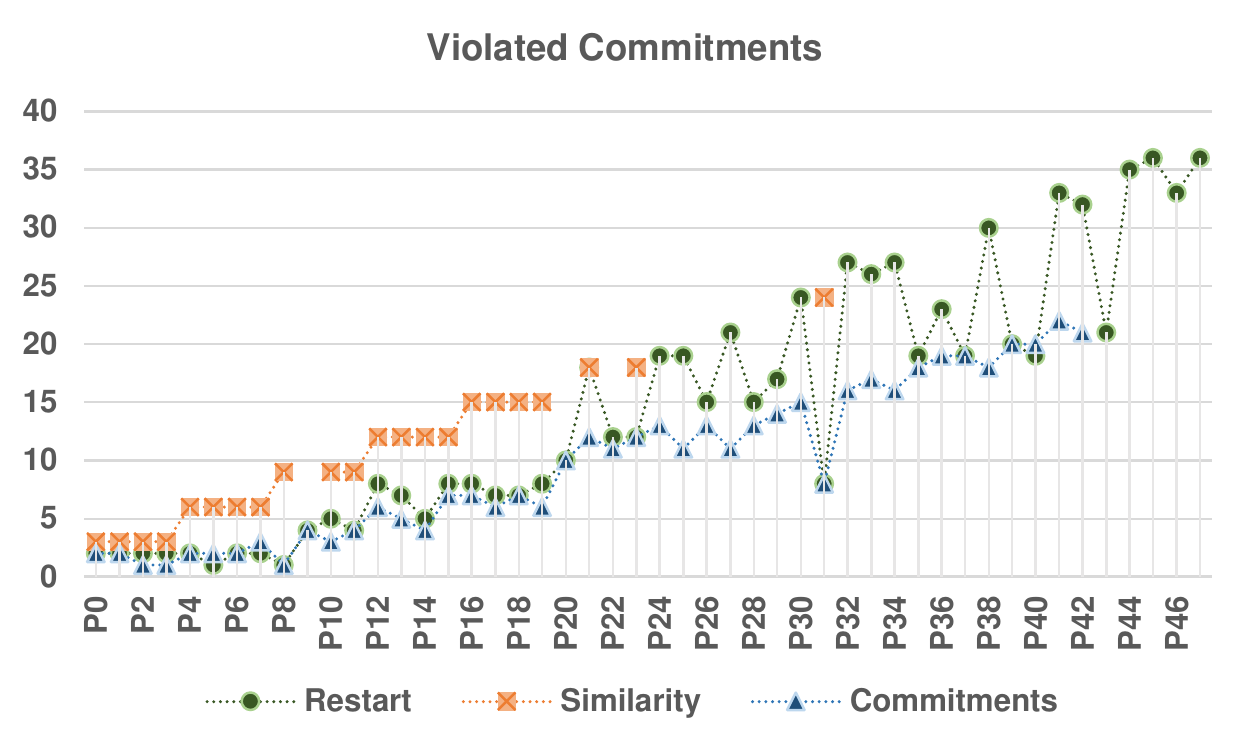}
  \vspace{-8mm}
  \caption{Number of agent commitments violated}
  \label{violated-comm}
\vspace{-3mm}
\end{figure}

\subsubsection{Metric: Commitment Satisfaction} 

Finally, we evaluated the number of inter-agent commitment violations in the
new plan, where the commitments come from the agent interactions in the
original plan. Figure~\ref{violated-comm} shows that the similarity preserving
method violates the most number of commitments in general. This may appear
surprising initially, since preserving the actions of the old plan are at
least tangentially related to preserving commitments between agents. However,
note that even the similarity maximizing method cannot return the \emph{exact
same} plan as the original one; some of the actions where it differs from the
old plan may indeed be the actions that created the inter-agent commitments in
the first place, while other preserved actions may now no longer fulfill the
commitments because the state of the world has changed. These results confirm
that both maximizing similarity as well as replanning from scratch are bad
surrogates for the metric of minimizing inter-agent commitment violations.






\section{Conclusion}

In this paper, we presented the idea that replanning ought to be looked at
less as a mere technique and more as a problem in its own right. We conducted
an overview of the various techniques that have been used as solutions to this
replanning problem, and the constraints on which they are based. We then
showed that the problems that these techniques solve can all be compiled into
a single substrate, as a means of comparing their effectiveness under
different planning metrics. After presenting this novel compilation, we showed
via an empirical evaluation that the various replanning techniques are not
good surrogates for each other. We thus focused a spotlight on the
incompatibility of the various replanning flavors with each other, due to the
disparate metrics that they seek to optimize.





\bibliographystyle{aaai}
\bibliography{paper}

\end{document}